\documentclass{article}

\usepackage{arxiv}

\usepackage{cite}
\usepackage{amsmath,amssymb,amsfonts}
\usepackage{algorithmic}
\usepackage{graphicx}
\usepackage{textcomp}
\usepackage{xcolor}
\def\BibTeX{{\rm B\kern-.05em{\sc i\kern-.025em b}\kern-.08em
    T\kern-.1667em\lower.7ex\hbox{E}\kern-.125emX}}

\usepackage[utf8]{inputenc} 
\usepackage[T1]{fontenc}    
\usepackage{hyperref}       

\usepackage{verbatim}       
\usepackage{adjustbox}      
\usepackage{multirow, tabularx, array, booktabs}        
\usepackage{rotating} 
\usepackage{listings} 

\title{Leaky ReLUs That Differ in Forward and Backward Pass Facilitate Activation Maximization in Deep Neural Networks}

\date{} 					

\author{ 
    \href{https://orcid.org/0000-0002-7039-5189}{\includegraphics[scale=0.06]{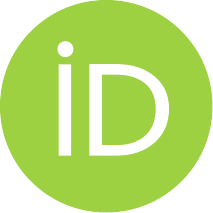}\hspace{1mm}Christoph Linse} \\
	Institute for Neuro- and Bioinformatics\\
	University of L{\"u}beck\\
	L{\"u}beck, Germany \\
	\texttt{c.linse@uni-luebeck.de} \\
	\And
	\href{https://orcid.org/000-0001-8556-2472}{\includegraphics[scale=0.06]{orcid.pdf}\hspace{1mm}Erhardt Barth} \\
	Institute for Neuro- and Bioinformatics\\
	University of L{\"u}beck\\
	L{\"u}beck, Germany \\
	\texttt{erhardt.barth@uni-luebeck.de} \\
    \And
	\href{https://orcid.org/0000-0002-4539-4475}{\includegraphics[scale=0.06]{orcid.pdf}\hspace{1mm}Thomas Martinetz} \\
	Institute for Neuro- and Bioinformatics\\
	University of L{\"u}beck\\
	L{\"u}beck, Germany \\
	\texttt{thomas.martinetz@uni-luebeck.de} \\
}

\renewcommand{\headeright}{}
\renewcommand{\undertitle}{}
\renewcommand{\shorttitle}{Leaky ReLUs That Differ in Forward and Backward Pass}

\hypersetup{
pdftitle={Leaky ReLUs That Differ in Forward and Backward Pass Facilitate Activation Maximization in Deep Neural Networks},
pdfauthor={Christoph Linse, Erhardt Barth, Thomas Martinetz},
pdfkeywords={convolutional neural networks, activation maximization, optimization, feature visualization, rectifiers},
}

\begin{document}
\maketitle

\begin{abstract}
Activation maximization (AM) strives to generate optimal input stimuli, revealing features that trigger high responses in trained deep neural networks. AM is an important method of explainable AI. We demonstrate that AM fails to produce optimal input stimuli for simple functions containing ReLUs or Leaky ReLUs, casting doubt on the practical usefulness of AM and the visual interpretation of the generated images. 
This paper proposes a solution based on using Leaky ReLUs with a high negative slope in the backward pass while keeping the original, usually zero, slope in the forward pass. The approach significantly increases the maxima found by AM.
The resulting ProxyGrad algorithm implements a novel optimization technique for neural networks that employs a secondary network as a proxy for gradient computation. This proxy network is designed to have a simpler loss landscape with fewer local maxima than the original network. Our chosen proxy network is an identical copy of the original network, including its weights, with distinct negative slopes in the Leaky ReLUs.
Moreover, we show that ProxyGrad can be used to train the weights of Convolutional Neural Networks for classification such that, on some of the tested benchmarks, they outperform traditional networks.
\end{abstract}


\section{Introduction}
\label{introduction}

Optimizing Convolutional Neural Networks (CNNs) that contain ReLUs \cite{nair_rectified_2010} or Leaky ReLUs \cite{maas2013rectifier} remains a central challenge in deep learning. The literature discusses optimization focusing on training the weights of deep neural networks \cite{douglas_why_2018, lu_dying_2020, whitaker_synaptic_2023}. While Leaky ReLU mitigates gradient sparsity, challenges persist in activation maximization (AM).

AM involves synthesizing input stimuli that maximize specific units within deep neural networks. A notable application involves optimizing an input image $\mathbf{x} \in [-1, 1]^{C \times H \times W}$ to activate a class neuron $i$ within the context of image classification. Here, we confine image intensity values to a range from -1 denoting black and 1 denoting white. Subsequently, the resulting image can undergo qualitative analysis to identify visual features that accurately depict the class.
AM can be described as finding
\begin{equation*}
    \mathbf{x}^* = \underset{\mathbf{x} \in [-1, 1]^{C \times H \times W}}{\text{argmax}} f_{s,i}(\mathbf{x})
\end{equation*}
with $f_{s,i}(\mathbf{x})$ being the output of the class neuron for class $i$. Here, $s$ is the negative slope of the Leaky ReLUs.
The optimization is achieved by gradient ascent
\begin{equation*}
    \mathbf{x} \longleftarrow R(\mathbf{x} + \mu \nabla_x f_{s,i}(\mathbf{x}))
\end{equation*}
with a learning rate $\mu$ and a regularization step $R$.

In the early stages of AM, Erhan et al. \cite{erhan_visualizing_2009} found that the maximum can be determined reliably and consistently for different randomly initialized input images. Modern CNNs have more layers, contain more parameters, use larger input images and prefer rectifiers over sigmoid activation functions.
We hypothesize that today's CNNs severely suffer from optimization issues during AM. We justify our hypothesis with what we call the "white image problem", an activation maximization problem where the solution is an input vector with maximal intensity values $\mathbf{x}^* = \mathbf{1}$. It consists of a function $\tilde{f}(\mathbf{x}) : [-1, 1]^{C \times H \times W} \longrightarrow [-1, 1]$ and a loss function $E(\mathbf{x}) = - \tilde{f}(\mathbf{x})$. A simple "white image problem" is given by $\tilde{f}(\mathbf{x}) = \sum \text{ReLU}(\mathbf{x})$ where the sum is over all input pixels. 
We show that AM does not find the optimal stimulus (white image) for simple functions containing ReLU or Leaky ReLU due to three optimization issues a) sparse gradients, b) the race of patterns, and c) local maxima. The issues a) - c) are explained in detail in Section \ref{sec:optimization_issues}.

It turns out that large negative slopes can mitigate these optimization issues. To achieve higher maxima during AM, we employ a secondary network as a proxy for gradient computation. This proxy network incorporates Leaky ReLUs with distinctively higher negative slopes, which leads to a simplified loss landscape with fewer local maxima compared to the original network, thereby enhancing optimization efficiency. The chosen proxy network is an identical copy of the original network, sharing the same weights.
In our ProxyGrad algorithm, the original network handles the forward pass, while the proxy network executes the backward pass. Notably, the latter considers the activations from the original network for gradient computation. This is different from simply replacing the ReLUs by Leaky ReLUs because the latter would employ Leaky ReLU in the forward and backward pass.
We demonstrate that class visualizations of ResNet18 \cite{he_deep_2016} trained on ImageNet \cite{russakovsky_imagenet_2015} reach higher activation values and seem to provide better visual clarity when compared to not using ProxyGrad.

The strong dependency of AM on the initialization of the input image could explain the success of techniques that use specific initialization images for AM. For instance, Nguyen et al. \cite{nguyen_multifaceted_2016} initialized the input with the average of an image cluster, a facet. Their technique successfully generates different aspects of the same class. 

Moreover, we show that ProxyGrad performs well when training the weights of CNNs for classification and, in some cases, outperforms traditional optimization. We train and test ResNet18 \cite{he_deep_2016} and PFNet18 \cite{linse_convolutional_2023} on three different datasets including the Caltech101 \cite{li_fei_fei_learning_2004}, Caltech-UCSD Birds-200-2011 \cite{wah_caltech_ucsd_2011}, and the 102 Category Flower dataset \cite{nilsback_automated_2008}.

The paper is structured as follows.
Section \ref{sec:related_work} summarizes the related work.
Section \ref{sec:optimization_issues} presents the optimization issues of ReLU and LeakyReLU concerning AM.
Subsequently, Section \ref{sec:ProxyGrad} introduces the ProxyGrad algorithm.
Section \ref{sec:activation_maximization} demonstrates the performance of ProxyGrad when visualizing classes from ImageNet.
In Section \ref{sec:recognition_performance}, we compare the recognition performances of ReLU, Leaky ReLU, and ProxyGrad for image classification.
Section \ref{sec:gradient_analysis} explains why ProxyGrad provides larger gradients with higher slopes within ResNet.
We discuss our results in Section \ref{sec:discussion}, specify the limitations of our work in Section \ref{sec:limitations}, and draw conclusions in Section \ref{sec:conclusion}.


\section{Related work}
\label{sec:related_work}

\subsection{Activation Maximization}

In 2009, Erhan et al. used AM to obtain qualitative interpretations of high-level features from neural networks trained on several vision datasets \cite{erhan_visualizing_2009}. Their goal was to gain insight into what a particular unit of a neural network represents.
AM plays a prominent role in explainable deep learning \cite{nguyen_understanding_2019, shahroudnejad_survey_2021, linse_convolutional_2023}.
Szegedy et al. \cite{szegedy_intriguing_2014} discovered that AM can produce adversarial examples. For initialization, they chose an image showing a specific class. Subsequently, they could slightly modify the image to activate another class using gradient information. A short transition vector in image space fooled the network to predict the other class while the visual difference was imperceptibly small to humans.
Nguyen et al. \cite{nguyen_deep_2015} generated artificial images that are unrecognizable for humans, but networks trained on ImageNet \cite{russakovsky_imagenet_2015} or MNIST \cite{lecun1998gradient} classified them as specific classes with high confidence.
Later, Linse et al. \cite{linse_walk_2022} used AM to visualize that some poorly generalizing CNNs are susceptible to high-frequency image patterns.
This paper discusses specific optimization issues during AM. We attribute these issues to the activation functions ReLU and Leaky ReLU.

\subsection{Activation Functions}
The ReLU \cite{nair_rectified_2010} produces sparse gradient vectors:
\begin{equation*}
\begin{split}
\text{ReLU}(x) &= \max (x, 0) \\
\frac{\partial \text{ReLU}(x)}{\partial x} &= (\text{ReLU}(x) > 0).
\end{split}
\end{equation*}
Here, the inequality expression is evaluated as 1 if correct, else as 0.
In order to improve the gradient flow, Leaky ReLU does not set the derivative to zero \cite{maas2013rectifier}:
\begin{equation*}
\frac{\partial \text{LReLU}_s(x)}{\partial x} = \max ((\text{LReLU}(x) > 0), s).
\end{equation*}
$s$ with $1>s>0$ is called the negative slope.

However, some visualization approaches modify the derivative of the ReLU. The Deconvnet approach \cite{fleet_visualizing_2014} applies the ReLU to the gradients, only allowing positive derivatives. The guided backpropagation approach \cite{springenberg_striving_2015} sets the derivatives with negative value or input to zero. The result is an even more sparse gradient vector.
In our approach, ProxyGrad modifies the derivatives by applying distinct negative slopes for Leaky ReLUs in the forward and backward pass. If the slope in the forward pass is zero and the slope of the backward pass is positive, the network produces sparse data representations, like the ReLU does, but dense gradients, like those obtained with the Leaky ReLU, in the backward pass.


\section{Optimization issues during activation maximization}
\label{sec:optimization_issues}

\begin{figure}[tb]
    \centering
    \adjustimage{width=0.3\linewidth}{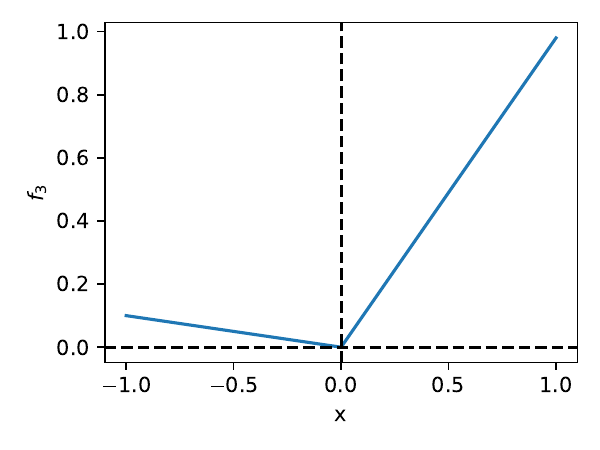}
    \caption{Illustration of function $f_3(\mathbf{x}) = \sum \text{LReLU}_s(\mathbf{x}) + \sum \text{LReLU}_s(-p\mathbf{x})$ with $p=0.2$ and $s=0.1$ for an input image with a single pixel.}
    \label{fig:proxygrad_f_3_plot}
\end{figure}

This section presents three optimization issues and demonstrates their relevance for AM using "white image problems" as toy examples. A white image problem is an AM problem where the optimal input pattern is $\mathbf{x}^* = \mathbf{1}$ (a white image). In the following, $\mathbf{x} \in [-1, 1]^{H \times W}$ is a gray scale image. The white image problem consists of a function $f(\mathbf{x}) : [-1, 1]^{H \times W} \rightarrow \mathbb{R}$ and a loss function $E(\mathbf{x}) = - f(\mathbf{x})$. Examples are shown in Table \ref{tab:optimization_issues}.

a) 'sparse gradients': The function $f_1(\mathbf{x}) = \sum \text{ReLU}(\mathbf{x})$ represents a white image problem where the sum is over all pixels. However, AM will not generate a white image for negative input values because of sparse gradients. This phenomenon is depicted in the second row of Table \ref{tab:optimization_issues}. Leaky ReLUs with a positive slope $s$ do not suffer from sparse gradients and generate the white image successfully.

b) 'race of patterns': The derivative of the Leaky ReLU implies that input elements are changed during gradient ascent at different speeds. The third row of Table \ref{tab:optimization_issues} shows the results of AM using the function $f_2(\mathbf{x}) = \sum \text{LReLU}_s(\mathbf{x})$. After few iterations, the positive values in the input increase their value, seen as white pixels. The negative values also increase, but at a much lower rate. Additional iterations are needed to produce the optimal input stimulus, the white image. In a deep CNN with many layers of Leaky ReLUs, the effect may significantly increase the iterations needed to reach local maximum.

c) 'local maxima': Some simple functions containing Leaky ReLUs have local maxima. The function $f_3(\mathbf{x}) = \sum \text{LReLU}_s(\mathbf{x}) + \text{LReLU}_s(-p \mathbf{x})$ with $1>p>s>0$ in Table \ref{tab:optimization_issues} represents a white image problem. As before, the sum is over the pixels. It consists of the Leaky ReLU plus another one scaled down by a factor $p$. The derivative after some specific pixel $x_i$ is $\frac{\partial f_3(\mathbf{x})}{\partial x_i}(1) = 1 - ps > 0$ and $\frac{\partial f_3(\mathbf{x})}{\partial x_i}(-1) = s - p < 0$, implying that gradient ascent will get stuck in a local maximum if the input contains negative values. Figure \ref{fig:proxygrad_f_3_plot} plots the function. One can see that the optimization will either end up in a black or a white pixel, depending on the initialization. Table \ref{tab:optimization_issues} shows an implementation of $f_3$ that applies two different weights on a single channel. No white image is generated. The optimization gets stuck in a local maximum.
Another example for local maxima uses a convolution operation with the kernel
{\footnotesize
$\left(\begin{smallmatrix}
    0 & 0 & 0 \\
    1 & 0 & -p \\
    0 & 0 & 0 
\end{smallmatrix}\right)$
}
as shown in Table \ref{tab:optimization_issues}. In the example, we chose the kernel specifically such that the network output is maximal when the input image is white (ignoring border effects). The large visual gap between the generated image and the optimal stimulus questions the practical usefulness of AM. Also, very different patterns are generated depending on the initialization of the input.

Nevertheless, as the negative slope $s$ increases toward 1, the race of patterns mends and the number of local maxima decreases. A slope of 1 leads to a linear network, making the AM problem convex. However, modifying the negative slope of a trained CNN also changes the predictions, adding to the difficulty of visually interpreting the visualization. Therefore, we propose to increase the slope in the backward pass only.

\begin{table}[tbp]
\caption{White image problems that are not solved via gradient ascent. A red (blue) box indicates positive (negative) values. The right column shows the result of AM. The initial image is shown at the top right. It is $1>p>s>0$ with the negative slope $s$.}
\label{tab:optimization_issues}
\centering
\begin{adjustbox}{width=0.8\linewidth}
    { \scriptsize
    \begin{tabular}{ccc}
    Function & Illustration & \adjustimage{width=1cm, valign=m, frame}{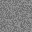} \\
    
    \midrule
    $\begin{array}{c}
    f_1(\mathbf{x}) = \\
    \sum \text{ReLU}(\mathbf{x}) \\
    \end{array}$
    & \adjustimage{trim=0.5cm 8.5cm 6.5cm 0.5cm, clip, width=1.3cm, valign=m}{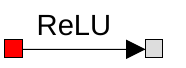}
    & \adjustimage{width=1cm, valign=m, frame}{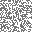} \\
    
    \midrule
    $\begin{array}{c}
    f_2(\mathbf{x}) = \\
    \sum \text{LReLU}(\mathbf{x}) \\
    \end{array}$ 
    & \adjustimage{trim=0.5cm 7cm 6.5cm 2cm, clip, width=1.3cm, valign=m}{local_maxima.pdf}
    & \adjustimage{width=1cm, valign=m, frame}{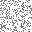}
    $\rightarrow$
    \adjustimage{width=1cm, valign=m, frame}{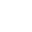} 
    \\
    
    \midrule
    $\begin{array}{c}
    f_3(\mathbf{x}) = \\
    \sum \text{LReLU}(\mathbf{x}) + \\
    \sum \text{LReLU}(-p\mathbf{x})\\
    \end{array}$
    & \adjustimage{trim=0.5cm 3.5cm 2cm 3.5cm, clip, width=3cm, valign=m}{local_maxima.pdf}
    & \adjustimage{width=1cm, valign=m, frame}{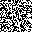} \\
    
    \midrule
    Convolution
    & \adjustimage{trim=0.5cm 0.5cm 0cm 7.4cm, clip, width=3.6cm, valign=m}{local_maxima.pdf}
    & \adjustimage{width=1cm, valign=m, frame}{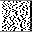} \\
    \end{tabular}
    } 
    \end{adjustbox}
\end{table}


\section{ProxyGrad}
\label{sec:ProxyGrad}

ProxyGrad employs a secondary network as a proxy for gradient computation. The proxy network should possess a simplified loss landscape with fewer local maxima than the original network, thereby enhancing optimization efficiency. Our proxy shares the architecture and the weights with the original network but incorporates a higher negative slope. During optimization, the loss is computed based on the original network. Subsequently, the gradient of the loss is computed based on the proxy but with the activations of the original network. Table \ref{tab:back_propagation} visualizes how ProxyGrad handles the backward pass with ReLUs in the original network and Leaky ReLUs in the proxy network. As seen in the example, the output remains sparse while the gradient becomes dense. The implementation of ProxyGrad is described in the Appendix \ref{sec:appendix:implementation}.

\begin{table}[tbp]
\caption{The forward pass (FP) and the backward pass (BP) for ReLU, Leaky ReLU, and ReLU with ProxyGrad. The gradient $R_i^l$ of the network output $f^{\text{out}}(\mathbf{x})$ after the $i$th feature map of layer $l$ is $R_i^l = \partial f^{\text{out}} / \partial f_i^l$.}
\label{tab:back_propagation}
\centering
\begin{adjustbox}{width=0.8\linewidth}
{ \scriptsize
\begin{tabular}{rcc}
FP ReLU with ProxyGrad: &
$f_i^{l+1} = \max(f_i^l, 0)$ &
\adjustimage{trim=2cm 21.5cm 2cm 2cm, clip, width=5cm, valign=m}{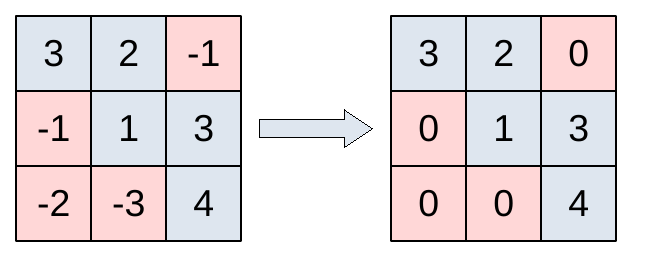} \\
BP ReLU: &
$R_i^l = (f_i^l > 0) \cdot R_i^{l+1}$ &
\adjustimage{trim=2cm 16.5cm 2cm 7cm, clip, width=5cm, valign=m}{gradient_example.pdf} \\
FP Leaky ReLU: &
$f_i^{l+1} = \max(f_i^l, s \cdot f_i^l)$ &
\adjustimage{trim=2cm 11.5cm 2cm 12cm, clip, width=5cm, valign=m}{gradient_example.pdf} \\
BP ReLU with ProxyGrad: &
$R_i^l = \max((f_i^l > 0), s) \cdot R_i^{l+1}$ &
\adjustimage{trim=2cm 6.3cm 2cm 17.2cm, clip, width=5cm, valign=m}{gradient_example.pdf} \\
\end{tabular}
} 
\end{adjustbox}
\end{table}


\section{Activation maximization using ProxyGrad}
\label{sec:activation_maximization}


\begin{table*}[bt]
\caption{Activation maximization of ResNet18 trained on ImageNet for a selection of classes. The negative slopes are 0.075. Best viewed digitally with zoom.}
\label{tab:fv_examples}
\centering
\begin{adjustbox}{width=\linewidth}
\setlength{\tabcolsep}{2pt}
{ \footnotesize
\begin{tabular}{cccc}
Class & ReLU & Leaky ReLU & ReLU with ProxyGrad \\ \addlinespace[4pt]

\toprule

Pelican &
\adjustimage{width=3cm, valign=m}{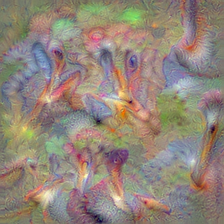} &
\adjustimage{width=3cm, valign=m}{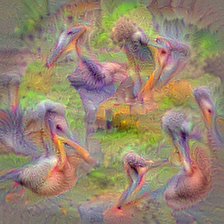} &
\adjustimage{width=3cm, valign=m}{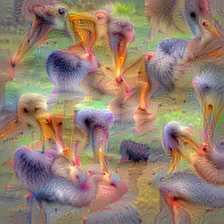}
\\ \addlinespace[4pt]

Lion &
\adjustimage{width=3cm, valign=m}{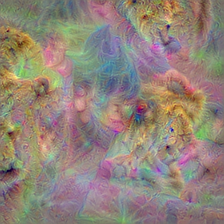} &
\adjustimage{width=3cm, valign=m}{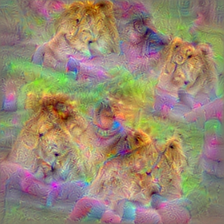} &
\adjustimage{width=3cm, valign=m}{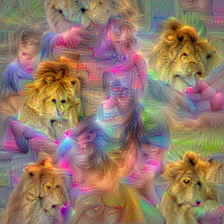}
\\ \addlinespace[4pt]

Analog cl. &
\adjustimage{width=3cm, valign=m}{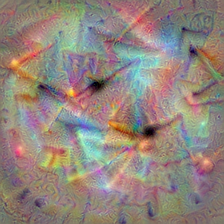} &
\adjustimage{width=3cm, valign=m}{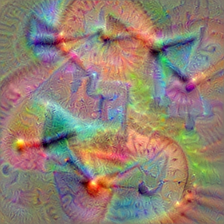} &
\adjustimage{width=3cm, valign=m}{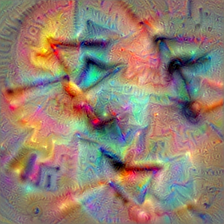}
\\ \addlinespace[4pt]

Bakery &
\adjustimage{width=3cm, valign=m}{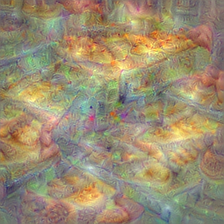} &
\adjustimage{width=3cm, valign=m}{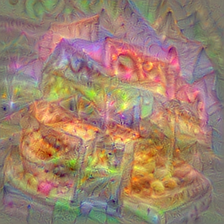} &
\adjustimage{width=3cm, valign=m}{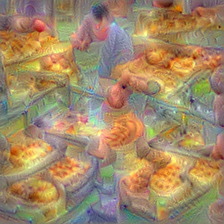}
\\
\end{tabular}

\begin{tabular}{cccc}

Class & ReLU & Leaky ReLU & ReLU with ProxyGrad \\ \addlinespace[4pt]

\toprule

Bucket &
\adjustimage{width=3cm, valign=m}{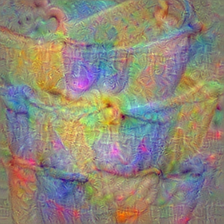} &
\adjustimage{width=3cm, valign=m}{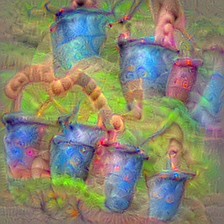} &
\adjustimage{width=3cm, valign=m}{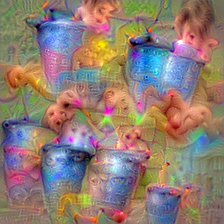}
\\ \addlinespace[4pt]

Poncho &
\adjustimage{width=3cm, valign=m}{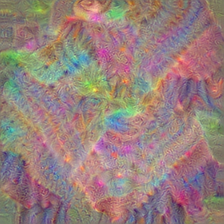} &
\adjustimage{width=3cm, valign=m}{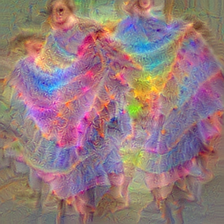} &
\adjustimage{width=3cm, valign=m}{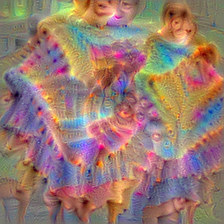}
\\ \addlinespace[4pt]

Bil. Table &
\adjustimage{width=3cm, valign=m}{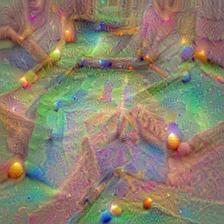} &
\adjustimage{width=3cm, valign=m}{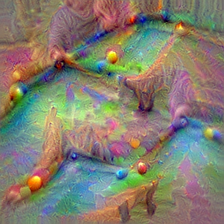} &
\adjustimage{width=3cm, valign=m}{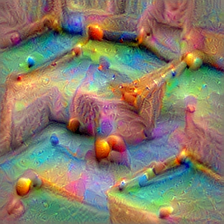}
\\ \addlinespace[4pt]

Cucumber &
\adjustimage{width=3cm, valign=m}{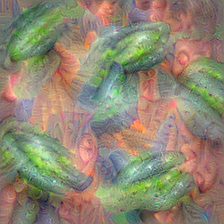} &
\adjustimage{width=3cm, valign=m}{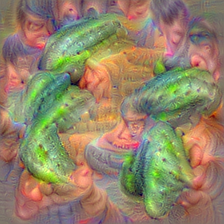} &
\adjustimage{width=3cm, valign=m}{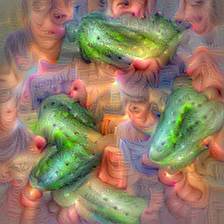}
\\

\end{tabular}
} 
\setlength{\tabcolsep}{6pt}
\end{adjustbox}
\end{table*}

To demonstrate the visualization capabilities of ProxyGrad, we perform AM on the class neurons of a ResNet18 pretrained on ImageNet \cite{russakovsky_imagenet_2015}. The original network contains ReLUs. The proxy uses Leaky ReLUs with a negative slope $s>0$ in the backward pass. We call this experiment ReLU with ProxyGrad. The experiments compare three methods: AM using the original ReLU network, AM using ReLU with ProxyGrad and AM using Leaky ReLU networks. In the latter experiment, we replace all ReLUs in the original network by Leaky ReLUs. All experiments use the same pretrained weights. The generated images are quantitatively evaluated using the class outputs of the original ReLU network. The initial image shows Gaussian noise on a gray background with $3 \times 224 \times 224$ pixels. The optimization lasts 10 or 500 iterations and uses a learning rate of $\mu = 25$. We either apply no regularization or weak image blur and small rotations after each iteration, a common strategy to find robust maxima \cite{yosinski_understanding_2015, shahroudnejad_survey_2021}. 
For comparability, the optimization uses normalized gradient vectors, as motivated by Section \ref{sec:gradient_analysis}, where the magnitude of gradients is discussed for different slopes. The update step becomes
\begin{equation*}
    x \longleftarrow R \left( x + \mu \frac{\nabla_x f(\mathbf{x})}{||\nabla_x f(\mathbf{x})||_2} \right).
\end{equation*}

\begin{figure}[tb]
	\centering
	\adjustimage{trim = 0.4cm 0.4cm 0.3cm 0.3cm, clip, width=0.48\linewidth}{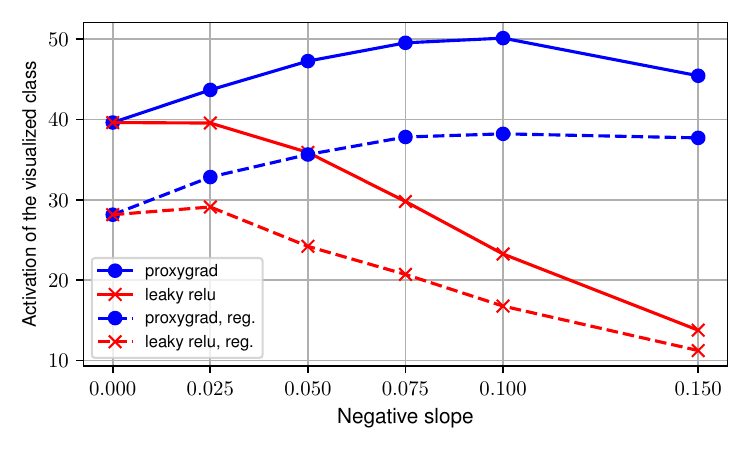}
	\adjustimage{trim = 0.4cm 0.4cm 0.3cm 0.3cm, clip, width=0.48\linewidth}{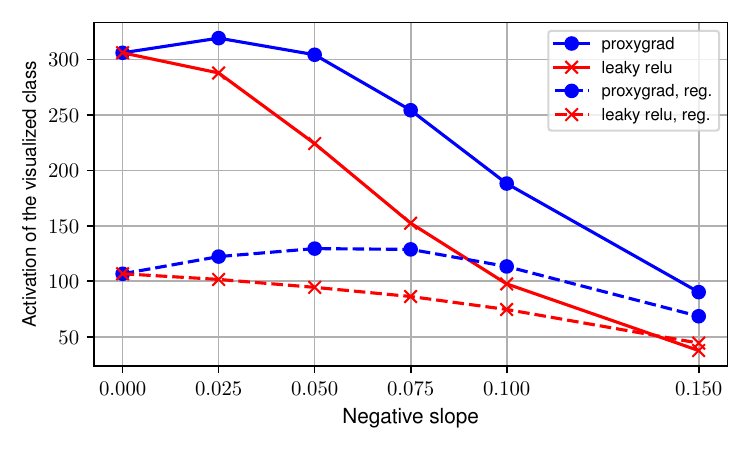}
	\caption{Mean activation of the generated images as measured by the original ReLU network. Top: after 10 iterations. Bottom: after 500 iterations.}
    \label{fig:activations_of_fv}
\end{figure}

Figure \ref{fig:activations_of_fv} presents the mean activation scores of the original ReLU network. The data point on the left hand side of the two diagrams ($s=0$) consider the images generated using the ReLU network (baseline). Images generated by ReLU with ProxyGrad are illustrated in blue and images generated by Leaky ReLU networks in red. 
ReLU with ProxyGrad reaches higher activation scores with increasing slope, showing its ability to facilitate optimization. Also, the ReLU network classifies the generated image as the target class in almost all cases.
After 10 iterations, images generated using ReLU with ProxyGrad have a mean activation of about 50 (40), while the images generated without ProxyGrad only reach 40 (30) without (with) regularization. After 500 iterations, ProxyGrad images have an activation of about 330 (130), while the images generated without ProxyGrad only have 300 (100) without (with) regularization. Because of gradient normalization, all optimization steps have equal size. Hence, the direction of the gradient returned by ProxyGrad is more stable during AM. 

However, for very high negative slopes, the activation scores start to decrease. Apparently, an optimal choice for the slope exists that might depend on many factors, including the regularization strategy. For the experiments with Leaky ReLU networks no such optimum seems to exist. For all slopes, the generated images tend to produce smaller activations than those generated by the original ReLU network.

For a qualitative comparison, Table \ref{tab:fv_examples} presents class visualizations for ReLU, Leaky ReLU and ReLU with ProxyGrad after 2500 iterations. The visual quality of the generated images tends to improve from left to right developing stronger key features of the respective class and more distinct shapes. The textures look cleaner and less noisy.

\begin{figure}[tb]
	\centering
	\adjustimage{width=0.25\linewidth}{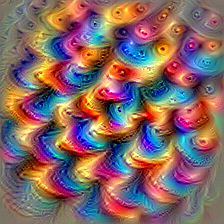}
	\adjustimage{width=0.25\linewidth}{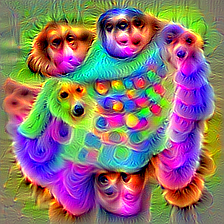}
	\caption{AM with a high negative slope of 0.3, far beyond the optimal range. In this negative example, the class 'poncho' is visualized. Best viewed digitally with zoom. Left: Leaky ReLU network. Right: ReLU with ProxyGrad}
    \label{fig:fv_very_high_slopes}
\end{figure}

Interestingly, for high negative slopes far beyond the optimal range, ReLU with ProxyGrad generates a kind of animal faces. This phenomenon is demonstrated in Figure \ref{fig:fv_very_high_slopes} where the class poncho is visualized. This occurs even if the class has nothing to do with animals or faces. We speculate that a dominant subnetwork that mainly processes facial features exists within the pretrained ResNet18. When performing AM using the Leaky ReLU network, very high negative slopes sometimes make the synthesized images collapse to simple, repetitive textures. Objects do not appear, as seen in Figure \ref{fig:fv_very_high_slopes}.

\section{Recognition performance}
\label{sec:recognition_performance}

This section addresses the performance of ProxyGrad for training the weights of CNNs on the Caltech101 dataset \cite{li_fei_fei_learning_2004}, the Caltech-UCSD Birds-200-2011 (CUB) dataset \cite{wah_caltech_ucsd_2011}, and the 102 Category Flower dataset (Flowers) \cite{nilsback_automated_2008}. The datasets are presented in more detail in Appendix \ref{sec:appendix:datasets}. In a comprehensive comparison, we train ReLU networks, ReLU networks with ProxyGrad, and Leaky ReLU networks.
The experiments are conducted for the ResNet18 architecture \cite{he_deep_2016} and the recently proposed Pre-defined Filter Network 18 (PFNet18) architecture \cite{linse_convolutional_2023}.
PFNet18 is a ResNet18 variant with depthwise convolution, where the spatial convolutional part does not change during training. The network reuses a fixed set of 16 different $3 \times 3$ edge kernels in the spatial convolutional part, which saves many parameters. Its learning process focuses entirely on learning linear combinations of pre-defined filter outputs. We chose PFNet18 because it follows a different learning philosophy and potentially exhibits different training dynamics. The networks are trained until convergence with the hyper-parameters shown in the Appendix \ref{sec:appendix:hyperparameters}.

\begin{table}[tb]
\caption{Average test performance of ResNet18 and PFNet18 on three benchmark datasets. $s$ denotes the negative slope.}
\label{tab:test_performance}
\centering
\begin{adjustbox}{width=0.4\linewidth}
\begin{tabular}{lllll}
\multicolumn{5}{c}{Caltech101 dataset} \\
\midrule
Architecture    & Activation & $s$ & Accuracy   & Std          \\
\midrule
\multirow{7}{5em}{ResNet18}    & ReLU with ProxyGrad     & 0.01           & 59.3  & 1.7  \\
    & ReLU with ProxyGrad     & 0.1            & 59.5  & 1.5 \\
    & ReLU with ProxyGrad     & 0.2            & \textbf{62.9}  & \textbf{0.8} \\
    & Leaky ReLU & 0.01           & 58.3  & 1.4  \\
    & Leaky ReLU & 0.1            & 59.3  & 1.6  \\
    & Leaky ReLU & 0.2            & 62.2  & 0.8  \\
    & ReLU       & 0              & 56.0  & 0.6  \\
\midrule
\multirow{7}{5em}{PFNet18}     & ReLU with ProxyGrad     & 0.01           & 63.5  & 1.0  \\
     & ReLU with ProxyGrad     & 0.1            & 63.7  & 0.7  \\
     & ReLU with ProxyGrad     & 0.2            & \textbf{64.6}  & \textbf{1.0}  \\
     & Leaky ReLU & 0.01           & 63.7  & 1.4  \\
     & Leaky ReLU & 0.1            & 63.4  & 0.9  \\
     & Leaky ReLU & 0.2            & 63.5  & 0.7  \\
     & ReLU       & 0              & 63.9  & 0.7  \\
\\ \addlinespace[4pt]

\multicolumn{5}{c}{CUB dataset} \\
\midrule
Architecture    & Activation & $s$ & Accuracy   & Std          \\
\midrule
\multirow{7}{5em}{ResNet18}    & ReLU with ProxyGrad     & 0.01           & 58.9  & 0.4  \\
       & ReLU with ProxyGrad     & 0.1            & 58.8  & 0.3 \\
       & ReLU with ProxyGrad     & 0.2            & 56.6  & 0.3 \\
       & Leaky ReLU & 0.01           & 59.6  & 0.3  \\
       & Leaky ReLU & 0.1            & \textbf{61.2}  & \textbf{0.4}   \\
       & Leaky ReLU & 0.2            & \textbf{61.2}  & \textbf{0.6}  \\
       & ReLU       & 0              & 58.8  & 0.5  \\
\midrule
\multirow{7}{5em}{PFNet18}     & ReLU with ProxyGrad     & 0.01           & 53.3  & 0.4 \\
        & ReLU with ProxyGrad     & 0.1            & 53.3  & 0.2 \\
        & ReLU with ProxyGrad     & 0.2            & 52.7  & 0.4  \\
        & Leaky ReLU & 0.01           & 53.3  & 0.3  \\
        & Leaky ReLU & 0.1            & 53.2  & 0.5\\
        & Leaky ReLU & 0.2            & \textbf{53.4}  & \textbf{0.3}  \\
        & ReLU       & 0              & 53.2  & 0.3  \\
\\ \addlinespace[4pt]

\multicolumn{5}{c}{Flowers dataset} \\
\midrule
Architecture    & Activation & $s$ & Accuracy   & Std          \\
\midrule
\multirow{7}{5em}{ResNet18}       & ReLU with ProxyGrad     & 0.01           & 74.1  & 0.2 \\
       & ReLU with ProxyGrad     & 0.1            & 74.7  & 0.5 \\
       & ReLU with ProxyGrad     & 0.2            & 75.3  & 0.4 \\
       & Leaky ReLU & 0.01           & 74.3  & 0.6 \\
       & Leaky ReLU & 0.1            & 75.7  & 0.4 \\
       & Leaky ReLU & 0.2            & \textbf{76.1}  & \textbf{0.3} \\ 
       & ReLU       & 0              & 74.0  & 0.3 \\
\midrule
\multirow{7}{5em}{PFNet18}        & ReLU with ProxyGrad     & 0.01           & 80.9  & 0.4 \\
        & ReLU with ProxyGrad     & 0.1            & 80.8  & 0.5 \\
        & ReLU with ProxyGrad     & 0.2            & 80.7  & 0.5 \\
        & Leaky ReLU & 0.01           & \textbf{81.0}  & \textbf{0.4} \\
        & Leaky ReLU & 0.1            & 80.9  & 0.4 \\
        & Leaky ReLU & 0.2            & \textbf{81.0}  & \textbf{0.4} \\ 
        & ReLU       & 0              & 80.8  & 0.2 \\
\end{tabular}
\end{adjustbox}
\end{table}

Table \ref{tab:test_performance} presents the mean test accuracy and the standard deviation of five experiments with different seeds. ReLU with ProxyGrad is well-suited for adjusting the weights of deep neural networks. ProxyGrad achieves the highest test accuracies for the Caltech101 dataset with both architectures. On the other two datasets, the Leaky ReLU is the best performer. However, the performance differences are minor.

High negative slopes between 0.1 and 0.2 tend to have the best test performance in our experiments. A negative slope of 0.01, the default setting used in many applications, belongs to the top performers only once (PFNet18 trained on the Flowers dataset). In the remaining five cases, a negative slope of 0.1 or higher reaches the highest test accuracies. However, higher slopes do not always perform better. We found that negative slopes higher than 0.2 lead to inferior performance.


\section{ProxyGrad provides large gradients}
\label{sec:gradient_analysis}

\begin{table}[tb]
\caption{Derivatives of the cross-entropy loss after the output of the first convolutional layer in the first block of ResNet18.}
\label{tab:outer_derivative}
\centering

\begin{tabular}{c}
Without batch normalization \\
\toprule
\adjustimage{width=0.45\linewidth, trim=0.3cm 0cm 0.3cm 0cm, clip, valign=m}{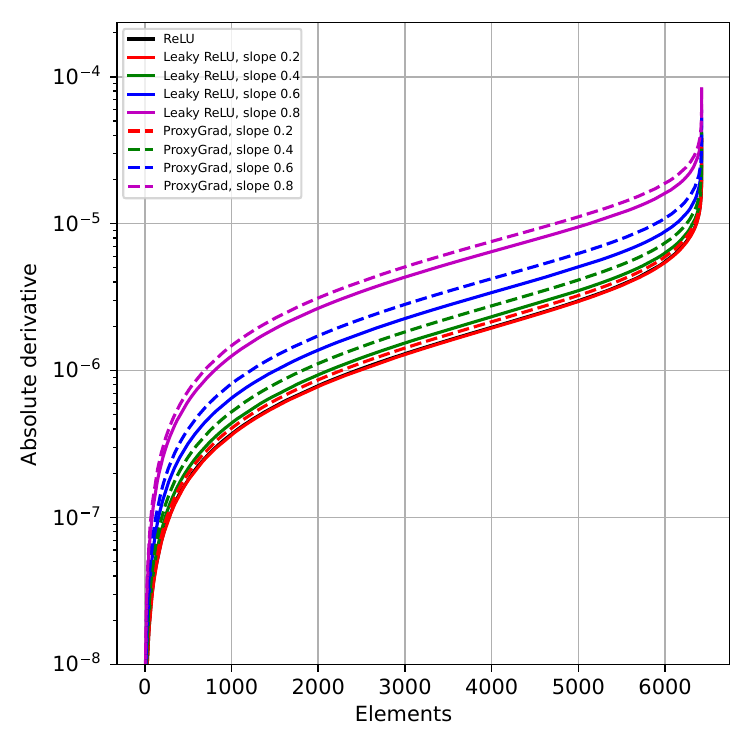} \\
\end{tabular}
\begin{tabular}{c}
With batch normalization \\
\toprule
\adjustimage{width=0.45\linewidth, trim=0.3cm 0cm 0.3cm 0cm, clip, valign=m}{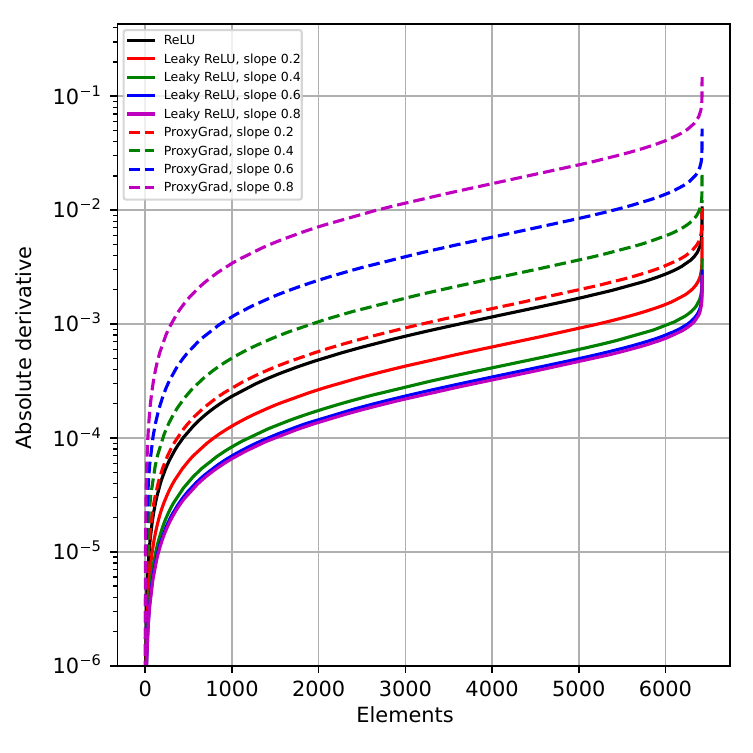} \\
\end{tabular}
\end{table}

We also study how the negative slope affects the magnitude of the gradients within ResNet18 in different layers. ReLU with ProxyGrad provides larger gradients with increasing slope. In Leaky ReLU networks, the gradients decrease with increasing slope. The latter might surprise, as the gradient magnitude of the Leaky ReLU alone increases with higher slope. However, we found that batch normalization effectively reduces the gradient. This finding has implications for the choice of hyper-parameters and supports our proposal to normalize the gradient in Section \ref{sec:activation_maximization}.

To study this phenomenon, we compute the derivatives of the cross-entropy loss in a randomly initialized ResNet18. For computation, we use randomly selected samples from the Caltech101 dataset. The experiment is repeated 50 times with different network initialization and input batches to obtain statistically robust results. Table \ref{tab:outer_derivative} shows the absolute derivatives after the output of the first convolutional layer in the first block. We observed similar results at different positions in the network. 
Without batch normalization, the derivatives increase with higher negative slope. When adding batch normalization layers, the derivatives decrease for Leaky ReLU networks. We attribute this phenomenon to the normalization step. Increasing the slope increases the variance of the output of a Leaky ReLU (see Appendix \ref{sec:appendix:rectified_gaussian}). The subsequent batch normalization layer divides the upstream gradient by the high standard deviation during the backward pass. Thus, the gradient magnitude decreases. This effect accumulates with every upstream batch normalization layer, leading to small gradients in the early layers within the network.
During training, the effect can be reduced by choosing a higher learning rate. ProxyGrad, however, does not suffer from gradient degradation and produces larger gradients.


\section{Discussion}
\label{sec:discussion}

AM is a challenging task considering networks with rectifiers. In contrast to training the weights of neural networks, where thousands or millions of train samples shape the loss landscape, AM relies on only one input image and one fixed set of weights. We showed that AM does not find the optimal stimuli for simple functions containing a ReLU or Leaky ReLU. This finding questions the practical application of AM and the visual interpretation of the generated images. Fortunately, the proposed ProxyGrad algorithm can enhance the maxima and the visual quality of the class visualizations. The resulting images show more clarity and structure. We attribute this improvement to the decreasing number of local maxima and the similar convergence speeds of different patterns.

Furthermore, ReLU networks with ProxyGrad and Leaky ReLU networks perform similarly on three challenging image classification tasks. Even though ProxyGrad does not follow exactly the gradient of the loss function, its gradient information is appropriate for decreasing the cross-entropy loss. 
ProxyGrad does not always offer superior classification performance, but provides some conceptual advantages over Leaky ReLUs. It can produce sparse output representations like a ReLU can while providing dense gradients. Also, a high negative slope in the backward pass of ProxyGrad does not lead to a linearization of the network, like Leaky ReLUs do.


\section{Limitations}
\label{sec:limitations}
This work considers CNNs with ReLUs and Leaky ReLUs because of their popularity in the Deep Learning community. Our discussion does not include other activation functions such as ELUs \cite{clevert_fast_2016}, SELU \cite{klambauer2017self}, Swish \cite{ramachandran_searching_2017}, or Mish \cite{misra_mish_2020}.
Also, this work does not study different optimizers for AM.

In our experiments, we observed an optimal range for the negative slope in the backward pass of ProxyGrad. If the negative slope is very high, far beyond the optimal range, AM can return low maxima and generate unplausible images. Therefore, we recommend users to determine the optimal range by trying out different negative slopes.


\section{Conclusion}
\label{sec:conclusion}

We identified three major optimization issues for activation maximization (AM) and deep networks with rectifiers: sparse gradients, the race of patterns, and local maxima. Toy examples showed that AM does not generate the optimal input stimuli even for simple functions containing a ReLU or Leaky ReLU. This result questions the practical application of AM and limits the visual interpretation of the generated images.
Nevertheless, a higher negative slope of the Leaky ReLU can mitigate these issues. To reach higher maxima, we proposed using different slopes in the forward and backward pass. Our ProxyGrad algorithm uses an identical copy of the original network with distinct negative slope as a proxy for gradient computation. ProxyGrad could successfully enhance the maxima returned by AM and the visual quality of the class visualizations using a ResNet18 pretrained on ImageNet. The generated images show more clarity and structure compared to using lower slopes. We attribute this improvement to the lesser number of local maxima and the simplification of the loss landscape.

We found that ProxyGrad is also well-suited for training the weights of deep neural networks. We trained and tested the ResNet18 and PFNet18 architectures on the Caltech101 \cite{li_fei_fei_learning_2004}, Caltech-UCSD Birds-200-2011 (CUB) \cite{wah_caltech_ucsd_2011} and the 102 Category Flower dataset \cite{nilsback_automated_2008}. Leaky ReLU and ProxyGrad performed similarly on the test sets. ProxyGrad is a good alternative to Leaky ReLUs because it benefits from the convergence properties of Leaky ReLUs while keeping the activation values sparse, as ReLUs do. In contrast to Leaky ReLUs, ProxyGrad does not tend to linearize the neural network with increasing negative slope.

As an outlook, the gradient flow could be further improved by moving the activation functions of ResNet to the convolutional branches such that the skip branches form a linear path from the input to the output layer of the network.
Another idea involves rethinking max and average pooling layers. In preliminary experiments we observed that different pooling methods strongly affect the generated patterns during AM. Analyzing these effects could lead to further insights and improvements of AM and network training.


\section*{Acknowledgment}
The work of Christoph Linse was supported by the Bundesministerium f{\"u}r Wirtschaft und Klimaschutz through the Mittelstand-Digital Zentrum Schleswig-Holstein Project.
The Version of Record of this contribution was presented at the 2024 International Joint Conference on Neural Networks (IJCNN),
and is available online at https://doi.org/10.1109/IJCNN60899.2024.10650881

\bibliographystyle{IEEEtran}
\bibliography{References}

\clearpage

\section{Appendix}

\subsection{Benchmark Datasets Used in Section \ref{sec:recognition_performance}}
\label{sec:appendix:datasets}
The Caltech101 dataset provides 8677 images and 101 classes including consumer products, instruments, plant species, animal species, and building types. In our experiments, we do not use the background class. There is no official test split available. For each seed, we randomly pick 20 images for training and 10 for testing per class.
The CUB dataset contains 5994 train and 5794 test images comprising 200 bird species. Often, the birds are in their natural habitats with plenty of clutter, foliage, tree branches, or flowers. Sometimes, the details of the birds cover tiny areas within the image. Furthermore, the bird categories have much intraclass variation in plumage color, pose, deformation, lighting and perspective.
The Flowers dataset contains 102 different blossom types that appear as one or multiple instances in each image. In contrast to the other datasets, the Flowers dataset has a train, validation, and test set. The official training and validation sets are merged into one training set. Hence, 2040 images exist in our training set and 6149 in the test set.

\subsection{Hyper-parameters for Network Training}
\label{sec:appendix:hyperparameters}
ResNet18 and PFNet18 are inititalized using Kaiming initialization \cite{he_delving_2015}. The Lamb optimizer \cite{you_large_2020} reduces the cross-entropy loss for 300 epochs with an initial learning rate of 0.003, a batch size of 64, and a weight decay of 1. The learning rate is stepwise reduced to $3 \cdot 10^{-6}$. The pre-processing divides the input images by the dataset's RGB mean and standard deviation. Augmentation steps enlarge the training set by random cropping and random horizontal flipping.

\subsection{Leaky ReLUs Applied on a Gaussian Distribution}
\label{sec:appendix:rectified_gaussian}

This section computes the mean and the variance of the output distribution of a Leaky ReLU applied on the mean-free Gaussian variable $X$ with variance 1.

\begin{equation*}
\begin{split}
   E(X^\text{LReLU}) &= \frac{1}{\sqrt{2\pi}} \int_{-\infty}^0 s x e^{-\frac{x^2}{2}} dx + \frac{1}{\sqrt{2\pi}} \int_0^{\infty} x e^{-\frac{x^2}{2}} dx \\
   &= -\frac{s}{\sqrt{2\pi}} + \frac{1}{\sqrt{2\pi}} = \frac{1-s}{\sqrt{2\pi}} \\
   V(X^\text{LReLU}) &= E({X^{\text{LReLU}}}^2) - E(X^\text{LReLU})^2 \\
   &= \frac{1}{\sqrt{2\pi}} \int_{-\infty}^0 s^2 x^2 e^{-\frac{x^2}{2}} dx + \frac{1}{\sqrt{2\pi}} \int_0^{\infty} x^2 e^{-\frac{x^2}{2}} dx \\
   & \quad - E(X^\text{LReLU})^2 \\
   &= \frac{s^2}{2} + 1 - \frac{1}{2} - \frac{(1-s)^2}{2\pi} = \frac{s^2 + 1}{2} - \frac{(1-s)^2}{2\pi} \\
\end{split}
\end{equation*}

The variance increases with increasing negative slope $s$. We used the following integrals where $\Phi(x)$ denotes the CDF of the Gaussian distribution:

\begin{equation*}
\int_a^b x e^{-\frac{x^2}{2}} dx = e^{-\frac{a^2}{2}} - e^{-\frac{b^2}{2}}
\end{equation*}
\begin{equation*}
\int_a^b x^2 e^{-\frac{x^2}{2}} dx = \sqrt{2\pi} (\Phi(b) - \Phi(a)) + (a e^{-\frac{a^2}{2}} - b e^{-\frac{a^2}{2}})
\end{equation*}

\begin{figure}[tb]
	\centering
	\adjustimage{width=0.6\linewidth}{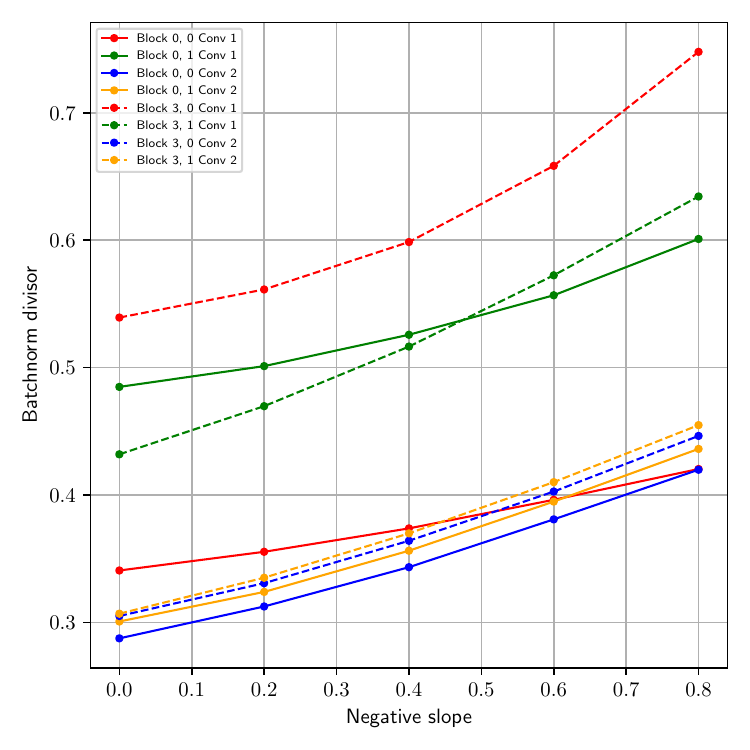}
        \caption{Standard deviation of the input to batch normalization layers taken from different locations in ResNet with Leaky ReLUs.}
    \label{fig:activations_out_plot}
\end{figure}

Furthermore, in ResNet18, we measure the standard deviation of the input to different batch normalization layers. We randomly initialize the network and select a random batch from the Caltech101 dataset as input. The experiment is repeated with different seeds to get statistically robust results. As shown in Figure \ref{fig:activations_out_plot} the standard deviation increases for higher negative slopes, which supports our hypothesis.

\subsection{Implementation of ProxyGrad in PyTorch}
\label{sec:appendix:implementation}

The following code uses ReLUs in the forward pass and Leaky ReLUs in the backward pass. To incorporate with existing code, replace all activation functions by ProxyGradRELU modules.

\begin{lstlisting}[language=Python, basicstyle=\ttfamily\footnotesize]
import torch
import torch.nn as nn
import torch.nn.functional as F

class ProxyGradRELUFN(torch.autograd.Function):
    @staticmethod
    def forward(ctx, x, negative_slope):
        ctx.x = x
        ctx.negative_slope = negative_slope
        return F.relu(x)

    @staticmethod
    def backward(ctx, grad_output):
        mask = ctx.x < 0.
        grad_output[mask] *= ctx.negative_slope
        return grad_output, None

class ProxyGradRELU(nn.Module):
    def __init__(self, negative_slope):
        super().__init__()
        self.negative_slope = negative_slope

    def forward(self, x):
        return ProxyGradRELUFN.apply(
            x, self.negative_slope)
\end{lstlisting}

\end{document}